# An EMD-based Method for the Detection of Power Transformer Faults with a Hierarchical Ensemble Classifier

Shoaib Meraj Sami[1] and Mohammed Imamul Hassan Bhuiyan
Department of Electrical and Electronic Engineering,
Bangladesh University of Engineering and Technology, Dhaka-1205, Bangladesh
Email: shoaib.eee08@gmail.com[1]

*Abstract*—In this paper, an Empirical Mode Decomposition-based method is proposed for the detection of transformer faults from Dissolve gas analysis (DGA) data. Ratio-based DGA parameters are ranked using their skewness. Optimal sets of intrinsic mode function coefficients are obtained from the ranked DGA parameters. A Hierarchical classification scheme employing XGBoost is presented for classifying the features to identify six different categories of transformer faults. Performance of the Proposed Method is studied for publicly available DGA data of 377 transformers. It is shown that the proposed method can yield more than 90% sensitivity and accuracy in the detection of transformer faults, a superior performance as compared to conventional methods as well as several existing machine learning-based techniques.

## I. INTRODUCTION

The reliability and stability of a power system including transmission and distribution considerably depends on the correct operation of its transformers. Faults in transformers can cause unscheduled outages of a particular area, device damages, explosion and economic loss. Further, explosion may put the lives of maintenance workers at risk. Obviously, minimization of these losses and accidents is an important problem and earlier fault detection and condition monitoring can be a great aid in this respect [1,6]. The transformer's insulation material, for example, mineral oil and cellulose paper condition can degrade and deteriorate because of electrical, mechanical, chemical and thermal stresses, ageing and abnormal working condition. Consequently, different types of faults such as partial discharge, arcing, sparking and thermal fault can occur. The deterioration of insulation material results in different types of gases, such as hydrogen ($H_2$), methane ($CH_4$), ethane ($C_2H_6$), ethylene ($C_2H_4$) acetylene ($C_2H_2$) carbon dioxide ($CO_2$) and carbon monoxide (CO) being dissolved in transformer oil. Thus, Dissolve gas analysis (DGA) analysis has become a valuable tool for the detection of faults and monitoring the condition of transformers [1,6]. Since its introduction, a variety of DGA-based techniques have been developed that include Duval triangle method, Dornenburg ratio method, Rogers ratio method, Key gas method and International Electro-Technical Commission Standard (IEC) code, among others [1-5]. Among these methods, Duval triangle method is widely used in many utility companies, but it poorly performs in the boundary between neighboring regions, and gives lower accuracy than the others when 'No result' cases are omitted. Dornenburg ratio method, Rogers ratio method, Key gas method and IEC code provide "No Result" in many cases [5]. Machine-learning-based techniques have been proposed in the literature to overcome the drawbacks of conventional fault diagnosis methods [6-11]. Many of these papers report moderate to satisfactory accuracy, leaving scope for further improvement. In recent times, empirical mode decomposition (EMD) has emerged as an effective tool for the analysis and processing of signals of a power system and its various components including transformers and rotating machines [12-14]. A major advantage of EMD over the Fourier and wavelet transform is that it does not need any prior basis function for decomposition. For example, by denoising acoustic and vibration signals, it is possible to better diagnose transformer winding deformation, winding overlapping, winding looseness, core deformation, partial discharge and rotary machine fault [12-14]. However, DGA-based transformer fault analysis is yet to be carried out in the EMD domain to the best of our knowledge. The objective of this paper is to introduce an EMD-based approach with machine learning and ratio-based DGA parameters for the detection of transformer faults with high accuracy. A simple method is proposed to rank a large set of DGA parameters using their skewness. Subsequently, the ranked parameters are decomposed in the EMD domain. Optimal sets of intrinsic mode function (IMF) coefficients are obtained and classified to detect the transformer faults. For this purpose, the IMFs of various sets of ranked features are classified with a single-layer XGBoost classifier. Consequently, the set of features yielding the highest accuracy is considered as the optimal set of features to be classified by a hierarchical classifier based on XGBoost. Experiments are carried out on publicly available DGA parameters of 377 transformers to study the performance of the proposed method and compare with those of several existing techniques.

## II. DESCRIPTION OF THE DGA DATA

In this Section, a brief description of the DGA data used in the paper is provided. The DGA parameters of 240 transformers from the Egyptian Electricity Network [15] along with those of the datasets used in [3,15] are utilized. Overall, the 377 transformers' DGA data are used in this work. It includes three different types of discharged faults, namely, partial discharge (PD), low energy discharge (D1), high energy discharge (D2), and three different types of thermal faults that are T1 (temperature less than $300^0$), T2 (temperature between $300^0$ to $700^0$) andT3 (temperature greater less than $700^0$). The distribution of different types of faults is illustrated in Table I.

Table I: Distribution of different fault types

| Fault Type | PD | D1 | D2 | T1 | T2 | T3 | Overall |
|---|---|---|---|---|---|---|---|
| Lab Samples | 27 | 42 | 55 | 70 | 18 | 28 | 240 |
| Literatures Samples | 15 | 25 | 59 | 10 | 3 | 25 | 137 |
| Total | 42 | 67 | 114 | 80 | 21 | 53 | 377 |

## III. METHODOLOGY

### A. DGA Parameters

Hydrogen ($H_2$), methane ($CH_4$), ethane ($C_2H_6$), ethylene ($C_2H_4$) and acetylene ($C_2H_2$) provide major information regarding fault of a transformer. The DGA-parameters are generally ratio type and used in conventional methods such as the Duval triangle method. These can be divided into: (i) dissolve gas concentration, (ii) gas ratio and (iii) gas relative percentage. Different parameters have different statistical discriminative power in different aspects. A total of thirty-seven parameters [8,15] are used in various works in the literature are listed in Table-II that are considered in this work for EMD-based classification and subsequent detection of transformer faults.

### B. Ranking of DGA Parameters and Optimal Feature Set

The feature selection and ranking method select a lower dimension but more informative features from the original set. The feature selection process is carried out in several steps:

#### Step 1: Ranking of DGA Parameters

The purpose of ranking the thirty-seven features is to generate a reliable and stable set of DGA features. In the literature, Student t-test, Kolmogorov-Smirnov test, Kullback-Leibler Divergence test, Pearson correlation coefficient, information gain, Fisher score, Laplacian score and ReliefF are used for ranking DGA based features [8-9]. In this paper, we use skewness for feature ranking because it is simpler and effective for representing intra-class compactness and inter-class separability between classes [16]. Table III shows that ranking of thirty-seven parameters from low to high skewness. Form the box plots in Fig.1, it is seen that for increasing the skewness, the distribution between classes becomes more asymmetric and ambiguous.

Table II: The DGA Parameters

| No. | Parameter | No. | Parameter |
|---|---|---|---|
| 1 | $H_2$/TH | 20 | THD |
| 2 | $CH_4$/TH | 21 | THH |
| 3 | $C_2H_6$/TH | 22 | TCH |
| 4 | $C_2H_4$/TH | 23 | $H_2$/THD |
| 5 | $C_2H_2$/TH | 24 | $CH_4$/THD |
| 6 | $C_2H_2/H_2$ | 25 | $C_2H_6$/THD |
| 7 | $C_2H_2/CH_4$ | 26 | $C_2H_4$/THD |
| 8 | $C_2H_2/C_2H_6$ | 27 | $C_2H_2$/THD |
| 9 | $C_2H_2/C_2H_4$ | 28 | $H_2$/THH |
| 10 | $C_2H_4/H_2$ | 29 | $CH_4$/THH |
| 11 | $C_2H_4/CH_4$ | 30 | $C_2H_6$/THH |
| 12 | $C_2H_4/C_2H_6$ | 31 | $C_2H_4$/THH |
| 13 | $(C_2H_4/H_2)+(C_2H_4/CH_4)$ | 32 | $C_2H_2$/THH |
| 14 | $H_2$ | 33 | $H_2$/TCH |
| 15 | $CH_4$ | 34 | $CH_4$/TCH |
| 16 | $C_2H_6$ | 35 | $C_2H_6$/TCH |
| 17 | $C_2H_4$ | 36 | $C_2H_4$/TCH |
| 18 | $C_2H_2$ | 37 | $C_2H_2$/TCH |
| 19 | TH | | |
| TH=$H_2$+$CH_4$+$C_2H_6$+$C_2H_4$+$C_2H_2$; THD=$CH_4$+$C_2H_4$+$C_2H_2$; THH=$H_2$+$C_2H_4$+$C_2H_2$; TCH=$CH_4$+$C_2H_6$+$C_2H_4$+$C_2H_2$ | | | |

#### Step 2: EMD-Based Features

The EMD is a mathematical method for extracting amplitude and frequency changing patterns from a given data. This method decomposes data or signal into integer sum of intrinsic mode function (IMF) [17-18]. The DGA parameters are ranked features are further subjected to empirical mode decomposition to elicit an optimal set of features. For this purpose, starting from lowest skewness of features that is 28, a set of twelve features are subjected to single level empirical mode decomposition. This operation is repeated for the next twenty-five sets of ranked features increasing skewness and so on. Thus, a set of twenty-six sets of intrinsic mode function are obtained. These twenty-six sets of IMF features are classified by using a single XGBoost classifier and corresponding values of accuracy are plotted in Fig.2. From the plot it is seen that the IMFs obtain from first ranked twelve DGA features (i.e. 28, 24, 1, 27, 31, 37, 26, 35, 36, 3, 32, 2) provides the best accuracy among the different IMF sets. Therefore, the IMF coefficients corresponding to the first twelve ranked DGA-based features are used in our final classification scheme that is presented in next section. Fig.3 shows the boxplots of the IMF coefficients for the six different classes of faults; every feature shows the discriminatory properties between different classes.

Table III: - Features by skewness (lower to higher)

| | Features Sorted by considering by higher to lower rank |
|---|---|
| Features No. from Table-II | 28, 24, 1, 27, 31, 37, 26, 35, 36, 3, 32, 2, 34, 4, 5, 33,21,14,19,20, 13, 10, 23, 6, 22, 15, 18, 17, 7, 8, 16, 11, 9, 12, 25, 30 & 29 |

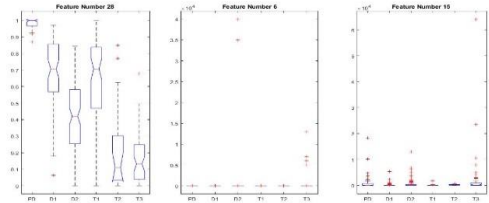

Fig. 1: Box plots of three DGA parameters (28, 6, 15 from Table-I) ranked by skewness

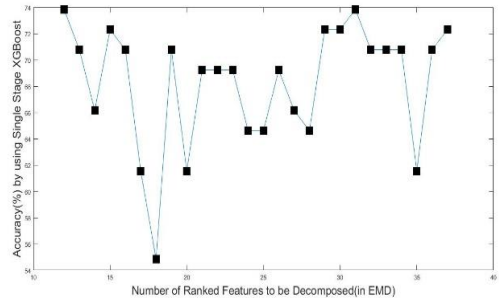

Fig. 2: Values of accuracy obtained by the single-stage XGBoost

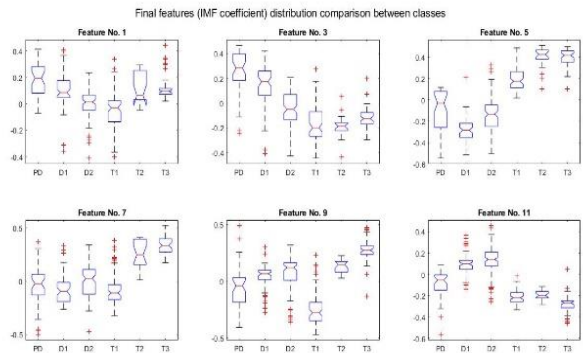

Fig. 3: Boxplots of the six IMF features (coefficients) for the six different classes of faults

### C. Hierarchical Classification

The proposed hierarchical classification scheme is presented. in Fig.4. As seen from the flowchart, basically the XGBoost classifiers are used in different levels of classification. First, an XGBoost classifier determines the discharge and thermal fault. Subsequently, two XGBoost classifiers distinguish specific fault from them in a similar fashion.

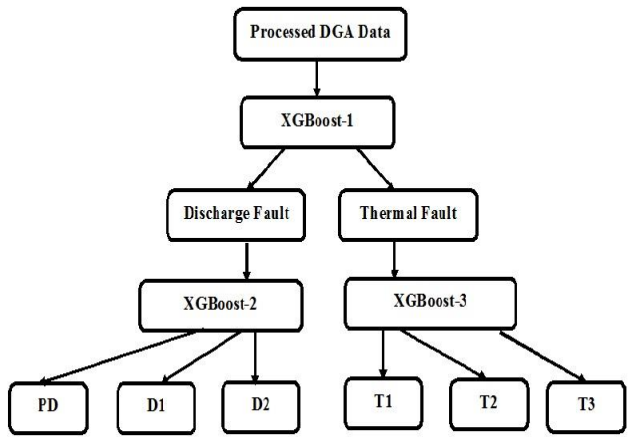

Fig. 4: Proposed Hierarchical Classification scheme

## IV. RESULTS AND DISCUSSION

In this Section, the performance of the proposed method is discussed. For each transformer from the 377 transformers dataset, a set of twelve IMF coefficients are obtained. Therefore, the overall size of the feature dataset considered for classification is 377×12. The features are classified by the hierarchical classification scheme in 85:15 ratios (training and testing, respectively). The experiments are conducted in Python environment (where the training and testing sets are selected randomly) on a Windows-10 64-bit platform having 4 GB RAM and 2.11 GHz Intel Core-i5 processor. For the purpose of comparison, well-known figures of merits such as accuracy and sensitivity are used. The performance of propose method is compared with existing results in literature including conventional methods and machine learning-based approaches.

Table IV provides the confusion matrix and sensitivity of all classes for the test set of features. The number of transformers in the test is 57 and shown in brackets for each class. Apart from the PD, the proposed method yields highly satisfactory values for sensitivity, and the average sensitivity is 91.12%.

In Table V, the performance of the proposed method is compared with that of other techniques. In general, the performances of the machine-learning techniques including the Proposed one give a superior performance as compared to the conventional methods. In addition, the proposed method yields an accuracy improvement of 7%, 14% and 28%, than the Ensemble Learning, HGA-SVM and BA-PNN-based approaches.

Table IV: Confusion matrix of the Proposed Method

| Predicted / Actual | PD | D1 | D2 | T1 | T2 | T3 | Sensitivity (%) |
|---|---|---|---|---|---|---|---|
| PD (7) | 5 | 2 | 0 | 0 | 0 | 0 | 71.43 |
| D1 (11) | 0 | 10 | 1 | 0 | 0 | 0 | 91.67 |
| D2 (19) | 0 | 1 | 18 | 0 | 0 | 0 | 94.73 |
| T1 (9) | 0 | 0 | 0 | 9 | 0 | 0 | 100 |
| T2 (1) | 0 | 0 | 0 | 0 | 1 | 0 | 100 |
| T3 (9) | 0 | 0 | 0 | 0 | 1 | 8 | 88.89 |

Table V: Comparison among various methods for the six fault classes

| Fault / Method | PD | D1 | D2 | T1 | T2 | T3 | Average Accuracy |
|---|---|---|---|---|---|---|---|
| Duval Method (%) | 14.28 | 0.5 | 68.42 | 55.56 | 0 | 88.89 | 57.89 |
| Rogers Four ratio method (%) | 0 | 0 | 78.95 | 100 | 100 | 33.33 | 49.12 |
| IEC method (%) | 28.57 | 33.33 | 52.63 | 77.78 | 0 | 55.56 | 66.67 |
| Ensemble Learning [11] (%) | 85.71 | 58.33 | 94.74 | 100 | 0 | 88.89 | 84.21 |
| BA-PNN [10] (%) | 85.71 | 41.67 | 73.68 | 77.78 | 100 | 33.33 | 63.16 |
| HGA-SVM [8] (%) | 71.43 | 66.67 | 94.73 | 44.45 | 0 | 100 | 77.19 |
| Proposed method (%) | 71.43 | 91.67 | 94.73 | 100 | 100 | 88.89 | 91.23 |

## V. CONCLUSION

A new machine-learning based technique is presented to detect transformer faults by classifying features extracted from DGA data. A total of thirty-seven ratio-based DGA parameters have been considered. These are analyzed in the EMD domain to extract an optimal set of IMF coefficients for representing the underlying fault of a transformer. The ability of the IMF coefficients in discriminating various fault classes has been demonstrated through box plots. In order to classify the features corresponding to a large set of transformers, a hierarchical classification scheme has been presented that employs XGBoost, a well-known ensemble classifier. A total of 377 transformers' DGA has been utilized that are publicly available and been used in the literature. The experimental results of the Proposed method shows that on average, it can achieve a high degree of sensitivity (91.12%) and accuracy (91.23%) in the detection of six classes of transformer faults, in general better than conventional techniques and a several machine-learning-based methods.